# Morphology-Aware Prognostic model for Five-Year Survival Prediction in Colorectal Cancer from H&E Whole Slide Images

Usama Sajjad[1*], Abdul Rehman Akbar[1], Ziyu Su[1], Deborah Knight[1], Wendy L. Frankel[1], Metin N. Gurcan[2], Wei Chen[1,] Muhammad Khalid Khan Niazi[1]

[1]Department of Pathology, The Ohio State University Wexner Medical Center, Columbus, OH, USA.

[2]Center for Artificial Intelligence Research, Wake Forest University School of Medicine, Winston-Salem, NC, USA.

**Correspondence**: Usama Sajjad, Email: Usama.Sajjad@osumc.edu

## Abstract

Colorectal cancer (CRC) remains the third most prevalent malignancy globally, with approximately 154,000 new cases and 54,000 projected deaths anticipated for 2025. The recent advancement of foundation models in computational pathology has been largely propelled by task agnostic methodologies that can overlook organ-specific crucial morphological patterns that represent distinct biological processes that can fundamentally influence tumor behavior, therapeutic response, and patient outcomes. The aim of this study is to develop a novel, interpretable AI model, PRISM (Prognostic Representation of Integrated Spatial Morphology), that incorporates a continuous variability spectrum within each distinct morphology to characterize phenotypic diversity and reflecting the principle that malignant transformation occurs through incremental evolutionary processes rather than abrupt phenotypic shifts.

PRISM is trained on 8.74 million histological images extracted from surgical resection specimens of 424 patients with stage III CRC. PRISM achieved superior prognostic performance for five-year OS (AUC = $0.70 \pm 0.04$; accuracy = $68.37\% \pm 4.75\%$; HR = 3.34, 95% CI = 2.28–4.90; $p < 0.0001$), outperforming existing CRC-specific methods by 15% and AI foundation models by ~23% accuracy. It showed sex-agnostic robustness (AUC $\Delta = 0.02$; accuracy $\Delta = 0.15\%$) and stable performance across clinicopathological subgroups, with minimal accuracy fluctuation ($\Delta = 1.44\%$) between 5FU/LV and CPT-11/5FU/LV regimens, replicating the Alliance cohort finding of no survival difference between treatments.

## Introduction

Colorectal cancer (CRC) is the third most widespread cancer, with approximately 154,000 new cases and 54,000 deaths estimated for 2025 [3]. This epidemiological burden is characterized by pronounced stage-dependent prognostic stratification, with 5-year survival rates of 94.7% for Stage I, 88.4% for Stage II, 74.3% for Stage III, and 31.5% for Stage IV [4]. CRC is morphologically heterogeneous, and current pathological assessment does not capture the morphological variability in a quantitative manner and lacks multi-variate (multi-feature) assessments. For example, there are associations between grade, necrosis, and stroma with CRC prognosis [5-8], illustrating that these underutilized morphological features, if analyzed together through a more quantitative and multi-variate approach, could provide deeper insights into patient outcomes and improve prognostic accuracy. Recently, Tumor budding (TB) has been recognized as markers of epithelial-mesenchymal transition and linked to poor prognosis [9]. Also, it has also been shown that necrosis is an independent prognostic variable with respect to progression-free survival [5]. Finally, stroma-rich tumors and those with immature desmoplastic stroma are also associated with worse outcomes [10]. These findings further underscore the prognostic value embedded in morphological characteristics—yet these morphological features remain underutilized in routine clinical assessments due to the lack of scalable, quantitative, and multi-variate analysis tools.

To address this gap, artificial intelligence (AI) systems have gained prominence, particularly with the digitization of tissue specimens into whole slide images (WSIs) [11, 12]. These technologies demonstrate robust performance in tumor classification, tumor subtyping, gene expression profiling, and automated analysis of nuclear/cellular features [12-17]. More recently, foundation models—trained using self-supervised learning on millions of WSIs —have emerged as a dominant paradigm in computational pathology [2, 18-21]. These models learn to identify generic visual patterns by minimizing feature distances between augmented versions of the same image (e.g., through color or rotation changes) [22]. However, this domain-agnostic training approach presents a critical limitation for CRC prognostication: it often overlooks organ-specific morphological features that reflect distinct biological processes and are essential for predicting tumor behavior, treatment response, and patient outcomes. As a result, despite their generalizability, these models may fall short in capturing the nuanced, morphology-driven prognostic signals crucial for CRC.

Given the morphological heterogeneity of CRC [23], a tool that fails to incorporate morphologically relevant heterogeneous features would be unable to adequately quantify and analyze the variability within distinct morphological regions [24]. While existing computational tools such as QuantCRC [24] have made significant contributions to CRC prognostication, they primarily focus on discrete quantification of morphological features without fully capturing the continuous spectrum of phenotypic variability present within distinct histological microenvironments within WSIs. This limitation is significant because cancer develops through gradual changes rather than sudden transformations. These continuous biological changes create a spectrum of cellular abnormalities rather than distinct categories, and these gradual morphological transitions directly influence tumor aggressiveness, metastatic potential, and therapeutic sensitivity. Therefore, an AI prognostic model is needed that integrates the diverse variability spectrum within each distinct morphology for prognostication.

To address these limitations, we propose **PRISM (Prognostic Representation of Integrated Spatial Morphology)**, an AI model that integrates CRC morphologically relevant heterogeneous features for five-year overall survival (OS) prognostication (Figure 1). PRISM advances beyond binary morphological detection and quantification by incorporating a continuous variability spectrum that characterizes the phenotypic diversity within each distinct morphological pattern. This approach captures subtle gradations in cellular architecture and tissue organization that exist within nominally similar morphological categories. Unlike pathologists who may label a region simply as "cancer," PRISM goes further by capturing the nuanced variations among neoplastic cells—for example, differences in nuclear morphology, gland formation, and architectural patterns—and leverages this detailed phenotypic diversity to enhance prognostication. PRISM starts by training morphology informed classifier and extracting the histological features of the following: (i) Epithelial grade Spectrum (Adenocarcinoma high grade, Adenocarcinoma low grade, Adenoma high grade, Adenoma low grade); (ii) Serrated Pathway Precursors (Sessile serrated lesion); (iii) Tumor Microenvironment (Stroma, Vessels, Inflammation, Necrosis); and (iv) Evidence of submucosal Invasion (Muscle). The PRISM captures biologically relevant information through a domain-specific branch and complements it with generic histopathological features extracted through a parallel branch, enabling a comprehensive and nuanced prognostic representation. This precise modeling of survival outcomes may enable more informed therapeutic decision-making, personalized treatment stratification, and improved resource allocation in clinical settings.

Our study also reveals a fundamental limitation in AI models development for prognosis: Conventional validation strategies such as K-fold cross-validation are demonstrably inadequate for robustly evaluating OS prediction models in histopathology. PRISM incorporates a comprehensive strategy that integrates clinical and pathological attributes prior to training and evaluation to minimize confounding effects. It also employs stratified evaluation across clinicopathological subgroups to ensure robust and equitable model performance. This rigorous approach enables PRISM to maintain prognostic accuracy across diverse clinical contexts and treatment scenarios.

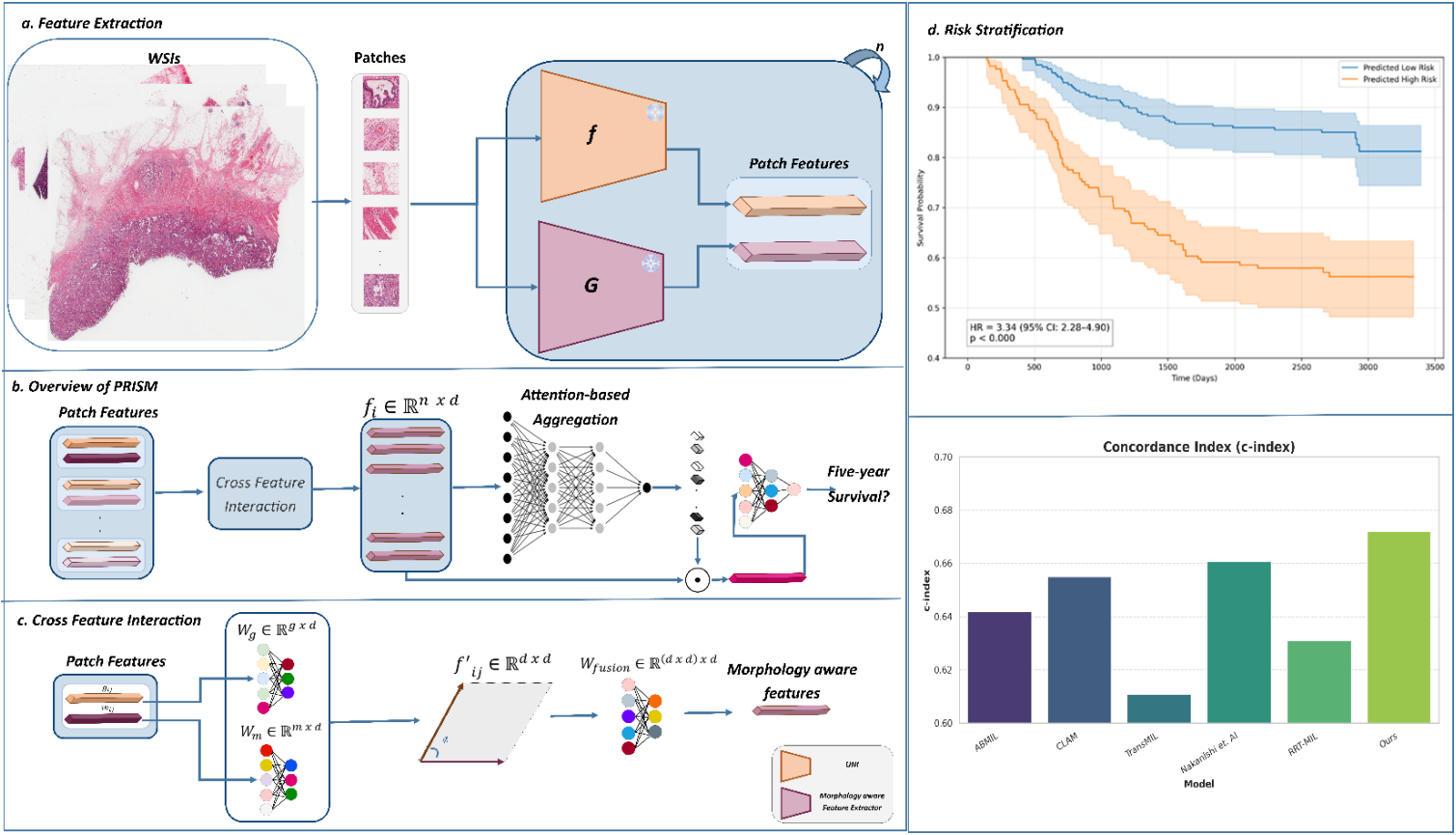

**Figure 1: An overview of our PRISM framework.** (a) We first tessellate WSIs into $n$ non-overlapping patches, with each patch undergoing dual feature extraction. (b) We perform cross-feature interaction between universal pathology features from UNI and morphology-aware features that encode tissue architecture and histopathological patterns. We then fuse these complementary feature representations $f_{i,j}$ at the patch level to create comprehensive morphological embeddings. An attention mechanism computes importance scores for each patch feature $f_{i,j}$ based on its prognostic relevance, enabling the model to focus on histologically relevant regions. We aggregate attention-weighted patch embeddings into a slide-level representation that captures the overall morphological landscape for five-year survival prediction. (c) During patch feature aggregation, we project features using two different neural networks ($W_g, W_m$) and aggregate the results to obtain morphology-aware features for each patch using $W_{fusion}$. (d) Based on the predicted probability, we train a time-to-event Cox Hazards model or perform risk stratification using concordance index.

## 2. Results:

### 2.1. Deep Learning Predicts Colorectal Morphological Phenotypes and Enables Extraction of Morphology Informed Features

The foundation of PRISM relies on accurate automated identification and quantification of distinct morphological phenotypes within colorectal histopathology. To establish this capability, we systematically validated our morphology informed phenotyping on comprehensively annotated tissue patches from HistAI colorectal dataset, ensuring robust performance across the diverse spectrum of morphological patterns encountered in CRC histopathology.

Our analysis demonstrated that training across this comprehensive spectrum of tissue morphologies yields a highly robust deep learning model for morphological phenotyping, achieving an overall accuracy of 90.43%, and AUC of 0.890. Critically, performance remained consistently high across all histological phenotypes, confirming the model's capacity to resolve the intrinsic morphological heterogeneity characteristics of CRC histopathology. This high-fidelity morphological phenotyping indicates that the learned feature representations encode biologically meaningful histopathological patterns. These resulting morphologically derived features capture essential organizational principles within the tumor microenvironment, enabling automated quantification of key histomorphometry parameters, including tumor burden, stromal composition, vascular density, and the spatial distribution of prognostically significant morphological components.

## 2.2. Prognostic Stratification of Stage III CRC Patients Using a Morphology-Aware Deep Learning Network

To evaluate the clinical utility and prognostic accuracy of PRISM, we conducted a comprehensive evaluation on the Alliance cohort [35]. Our assessment focused on quantifying improvements in survival prediction accuracy while examining the model's ability to maintain balanced performance across key clinical metrics. We compared PRISM against state-of-the-art multiple instance learning approaches (CLAM, TransMIL, ABMIL, RRT-MIL, Nakanishi et. al) [1, 29-32] and existing CRC prognostication methods to establish its relative performance and clinical significance. Additionally, we conducted detailed analysis of sensitivity-specificity trade-offs and cross-validation stability to assess the PRISM's readiness.

First, we applied PRISM on the Alliance cohort (N=424 patients), demonstrating significantly enhanced prognostic stratification for five-year post-resection survival (Figure 2). Overall, PRISM achieved an AUC of $0.70 \pm 0.04$ and an accuracy of $68.37\% \pm 4.75\%$, demonstrating >9% improvement in accuracy over existing approaches (second-best: $58.84\% \pm 4.50\%$). We also compared PRISM with Nakanishi et. al [32] that was proposed specifically for predicting CRC recurrence and demonstrated an improvement of 9% over their approach. Additionally, PRISM's superior performance was statistically significant compared to all baseline methods (p-values: ABMIL = 0.012, CLAM = 0.018, TransMIL < 0.0001, Nakanishi et al. < 0.0001, RRTMIL < 0.0003).

We also evaluated PRISM's performance using both sensitivity (accurate identification of patients who died within five years) and specificity (accurate detection of survivors at five-years). As shown in Figure 2, PRISM achieved an optimal balance between these metrics, while comparison methods (CLAM [29], TransMIL [30], Ilse et al. [1], RRT-MIL [31], Nakanishi et al. [32]) exhibited 10-20% difference between sensitivity and specificity. While threshold adjustment could improve this balance for individual method, it inevitably compromises the opposing metric. The consistently lower standard deviation of our model compared to alternatives demonstrates superior robustness, highlighting the importance of incorporating relevant histopathological features.

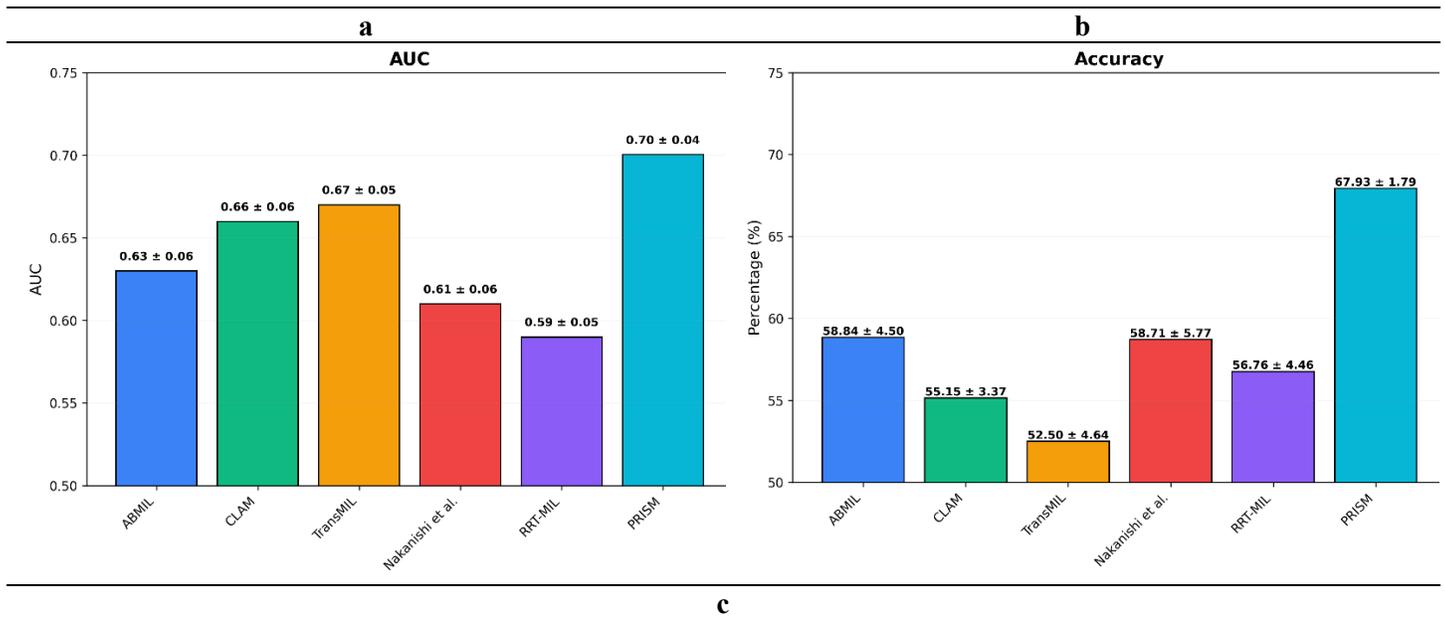

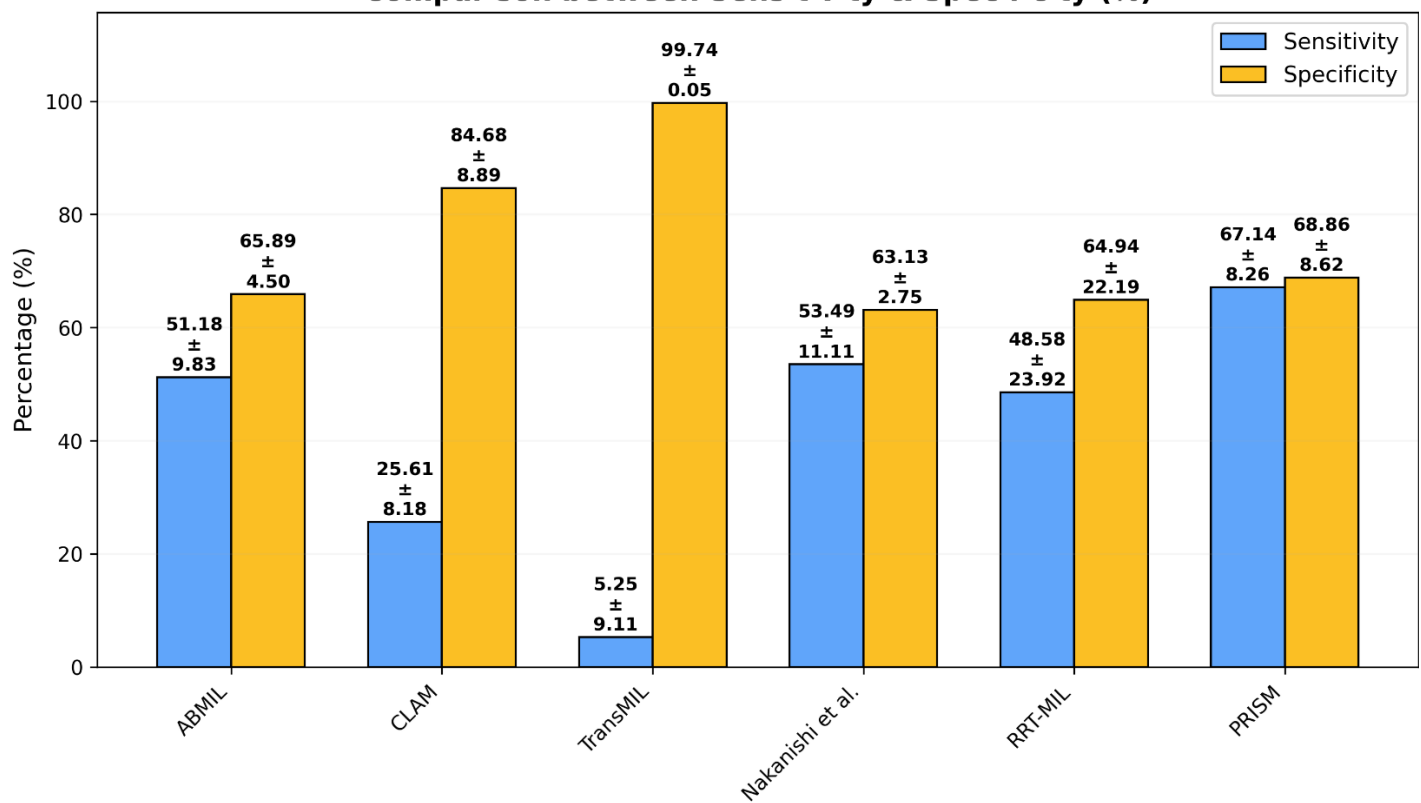

**Figure 2:** Five-year survival prediction results in the Alliance cohort using five-fold cross-validation. (a) Area Under the Curve (AUC) values with standard deviations, (b) Accuracy percentages with standard deviations, and (c) Grouped comparison of sensitivity and specificity with other models. Our model achieves the highest accuracy and balanced sensitivity and specificity, demonstrating superior performance compared to existing state-of-the-art methods (CLAM [29], TransMIL [30], ABMIL [1], RRT-MIL [31], Nakanishi et al. [32]).

## 2.3. Comparison of Five-Year Survival Prediction Using PRISM Across Different Treatments

The potential impact of different chemotherapeutic regimens on tumor biology necessitates examining prognostic model performance across treatment-specific cohorts. While the original Alliance trial found no significant survival difference between FL and IFL treatment regimens, variations in model performance could indicate irrelevant learned features rather than intrinsic biological features.

We compared the performance of benchmarks methods with PRISM by stratifying the treatment (FL vs IFL). The results are presented in Table 1 respectively. PRISM demonstrated robust performance with the highest AUC (0.7160 ± 0.061) and accuracy (68.21% ± 3.40), maintaining balanced sensitivity-specificity (64.82% ± 3.40 / 71.60% ± 8.50) on the Alliance cohort treated with FL only. In contrast, benchmark models exhibited critical limitations: TransMIL showed sensitivity collapse, and ABMIL's, Nakanishi et. al [32], RRT-MIL's high sensitivity, and accuracy volatility (±20.00 SD, ±11.43 SD) revealed fundamental instability to learn the morphologically meaningful features. Transitioning to the Alliance cohort treated using IFL treatment, PRISM maintained a consistent accuracy (66.77% ± 5.10) and competitive AUC (0.6846 ± 0.108) with a degradation of ~0.030 AUC and 1.45% in accuracy across FL, IFL treatment groups, whereas benchmarks displayed high performance degradation; ABMIL's AUC decreased by ~7%, CLAM's accuracy dropped 10%, Nakanishi et Al. [32] accuracy differs 5% across FL/IFL treatment groups, and RRT-MIL's variability to accurately predict patients survived five-years surged to ± 25.00 SD.

**Table 1:** Five-year survival prediction results of PRISM stratified by FL/IFL treatments in the Alliance cohort using five-fold cross-validation. Each row reports the average performance with standard deviation.

| Model | Treatment | AUC | Accuracy (%) | Sensitivity (%) | Specificity (%) |
|---|---|---|---|---|---|
| ABMIL [29] | FL | 0.65 ± 0.11 | 62.02 ± 13.76 | 51.93 ± 22.92 | 72.12 ± 08.70 |
|  | IFL | 0.58 ± 0.07 | 54.13 ± 05.61 | 45.34 ± 08.90 | 62.92 ± 06.89 |
| CLAM [1] | FL | 0.61 ± 0.09 | 50.04 ± 05.96 | 14.12 ± 08.97 | 85.96 ± 08.47 |
|  | IFL | 0.63 ± 0.15 | 60.89 ± 06.01 | 36.71 ± 13.97 | 85.07 ± 09.81 |
| TransMIL [30] | FL | 0.75 ± 0.10 | 54.75 ± 08.37 | 10.33 ± 16.13 | 99.16 ± 01.66 |
|  | IFL | 0.62 ± 0.09 | 51.81 ± 03.63 | 3.63 ± 07.27 | 100.00 ± 00.00 |
| Nakanishi et. Al [32] | FL | 0.65 ± 0.16 | 61.12 ± 12.85 | 48.38 ± 21.23 | 73.85 ± 07.18 |
|  | IFL | 0.55 ± 0.07 | 56.02 ± 05.29 | 56.23 ± 10.32 | 55.80 ± 08.49 |
| RRT-MIL [31] | FL | 0.59 ± 0.15 | 55.42 ± 08.54 | 43.45 ± 25.65 | 67.38 ± 25.52 |
|  | IFL | 0.55 ± 0.04 | 55.62 ± 03.61 | 46.90 ± 24.09 | 64.35 ± 25.62 |
| PRISM | FL | 0.72 ± 0.06 | 68.21 ± 03.40 | 64.82 ± 03.40 | 71.60 ± 08.50 |
|  | IFL | 0.68 ± 0.11 | 66.77 ± 05.10 | 68.76 ± 13.90 | 64.77 ± 10.90 |

### 2.4. Morphology-Informed Features Correlates with Time-to-Event Survival

Beyond predicting five-year classification, we evaluated PRISM's capacity for continuous risk stratification and survival prediction using established survival analysis metrics. This assessment provides critical insights into PRISM's clinical utility for patient counseling and treatment planning decisions.

PRISM demonstrated remarkable performance in stratifying CRC patients into distinct prognostic risk categories using Cox proportional-hazards model, achieving a hazard ratio of 3.34 (95% CI: 2.28–4.90) and concordance index (c-index) of 0.67 (Figure 4; Supplementary Table S6). This represents a significant improvement over all benchmark methods: CLAM (HR: 1.96, 95% CI: 1.30–2.98; c-index: 0.62), Nakanishi et al. (HR: 1.84, 95% CI: 1.27–2.65; c-index: 0.60), ABMIL (HR: 1.67, 95% CI: 1.16–2.41; c-index: 0.61), TransMIL (HR: 6.30, 95% CI: 3.05–13.03; c-index: 0.65), and RRT-MIL (HR: 1.47, 95% CI: 1.01–2.12; c-index: 0.56). While TransMIL showed a higher nominal hazard ratio, its broad confidence interval and reduced accuracy limits its ability for individualized risk stratification. Additionally, we visualized the Kaplan–Meier survival curves overall survival probability for five-year follow-up in Figure 3. PRISM achieved the highest hazard ratio of 3.34 (95% CI: 2.28-4.90) and concordance index of 0.67. While benchmark methods such as CLAM [29] and Nakanishi et al. [32] show initial separation between risk groups, their curves demonstrate earlier convergence and less pronounced differences in survival probabilities, with ABMIL [1] displaying moderate separation but notable fluctuations suggesting less stable risk predictions, TransMIL [30] exhibiting erratic curve behavior with wide confidence intervals despite a high nominal hazard ratio (6.30), and RRT-MIL [31] demonstrating the poorest performance with minimal curve separation and the lowest hazard ratio (1.47).

**ABMIL** **CLAM**

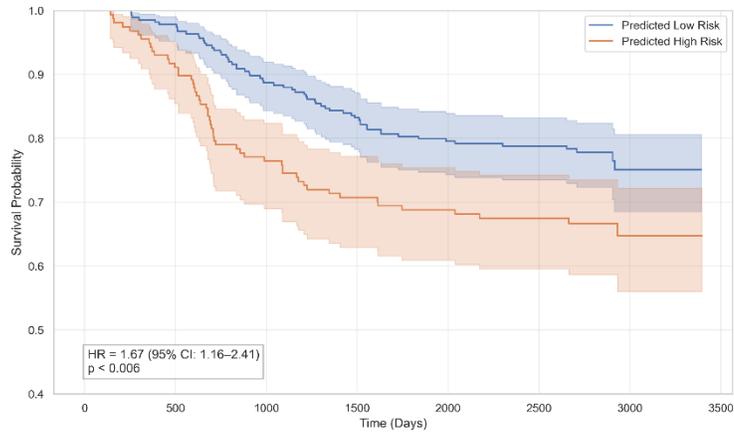 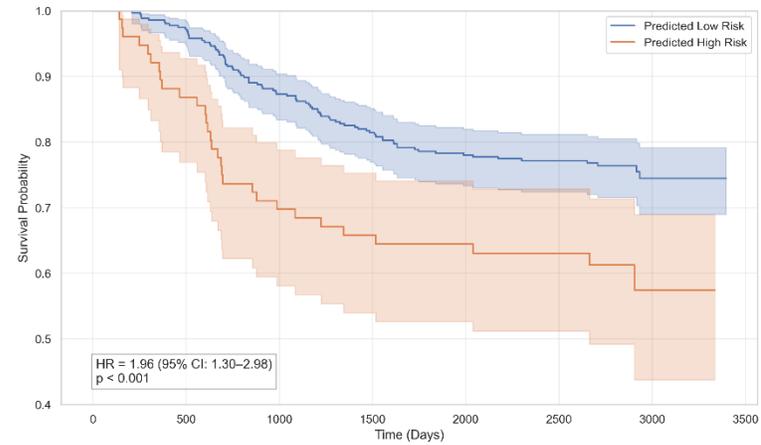

**TransMIL** **Nakanishi et. Al**

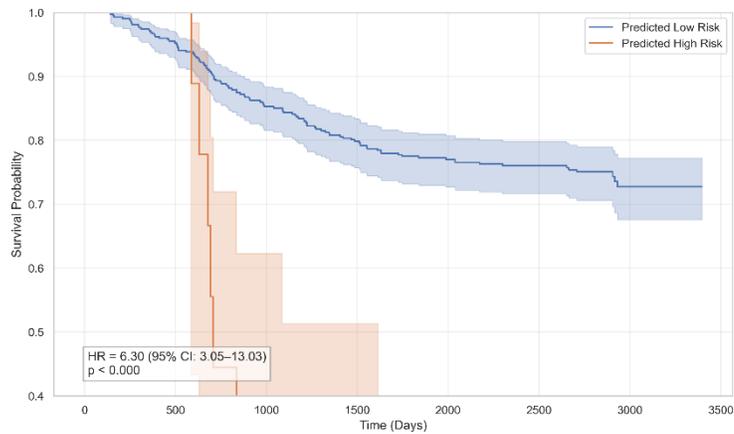 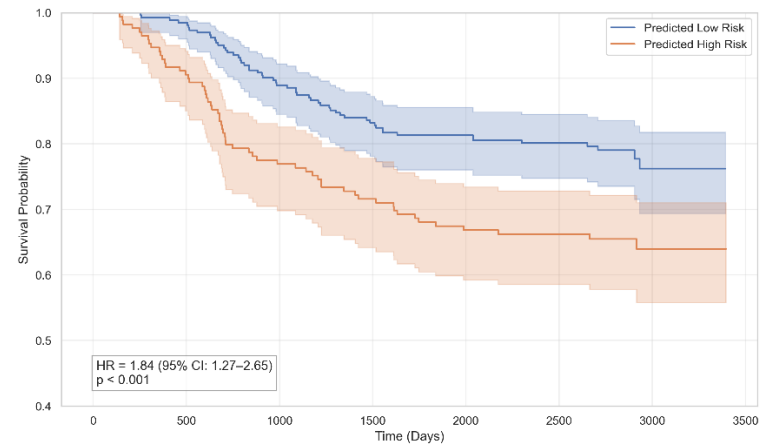

**RRT-MIL** **PRISM**

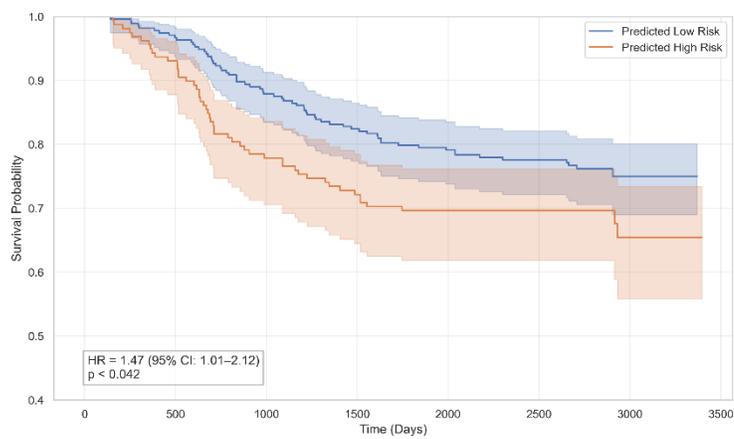 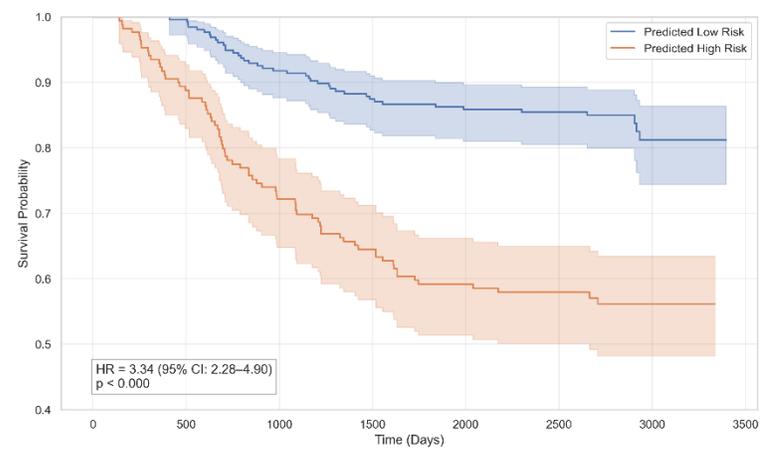

**Figure 3.** Kaplan–Meier plots for overall survival probability for five-year follow-up.

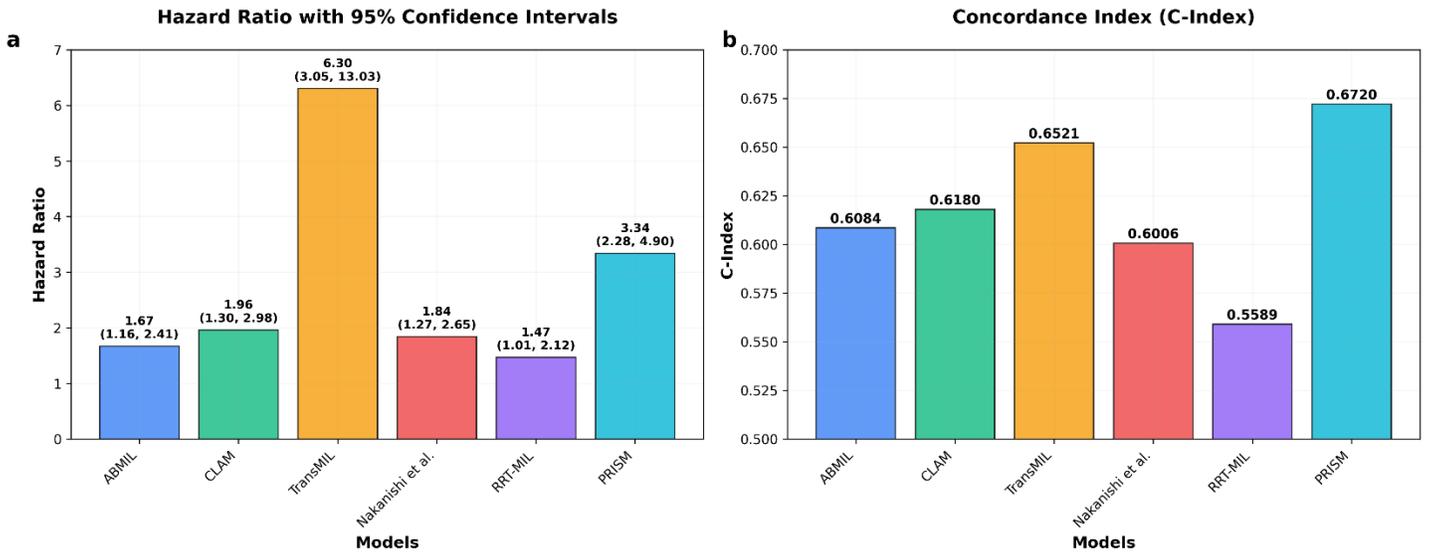

**Figure 4:** Hazard ratios and concordance index (c-index) values in the Alliance cohort using five-fold cross-validation. (a) Hazard ratios with 95% confidence intervals for each model, and (b) Concordance Index (C-Index) values demonstrating model ability to sort the patient based on risk for survival prediction. Our model achieves the highest C-Index (0.6720) and comparable hazard ratio in comparison with TransMIL [30] with smaller confidence interval, indicating superior prognostic performance compared to other state-of-the-art methods.

## 3. Discussion

This study presents PRISM, a novel morphology-informed AI prognostic model that significantly advances the field of AI-driven prognostication in CRC. Our findings demonstrate that incorporating specific morphological features into deep learning models yields substantial improvements over conventional approaches, while simultaneously revealing critical limitations in current model validation practices and highlighting the importance of addressing algorithmic bias in clinical AI applications (Supplementary Results 7.1-7.3).

### 3.1. Enhanced Prognostic Performance Through Novel, Morphology-Aware Feature Integration

The superior performance of PRISM, achieving an AUC of 0.70 ± 0.04 and accuracy of 68.37% ± 4.75%, represents a meaningful advance over existing methodologies. The ~9% improvement in accuracy over the second-best performing method translates to more precise risk stratification for approximately 1 in 10 patients, which could significantly impact clinical decision-making at population scales. More importantly, the hazard ratio of 3.34 (95% CI: 2.28–4.90) demonstrates robust discrimination between high- and low-risk patient populations, providing clinicians with actionable prognostic information that extends beyond traditional TNM staging.

The success of PRISM validates the hypothesis that CRC's inherent morphological heterogeneity contains prognostically relevant information that is inadequately captured by conventional histopathological evaluation. By explicitly training on the Hist-AI colorectal dataset [25] to recognize specific tissue morphologies that includes glandular carcinogenesis spectrum, serrated pathway precursors, tumor microenvironment characteristics, and invasion biomarkers, our model learns biologically meaningful features that align with established pathological knowledge. This targeted feature extraction addresses a fundamental limitation in foundation model pretraining, where self-supervised objectives are decoupled from the morphological patterns most relevant for prognostication.

Our findings also reproduce the Alliance cohort clinical trial results, confirming no significant survival differences between treatment groups (5FU/LV versus CPT-11/5FU/LV). These treatment-driven performance differences highlight critical flaws in benchmark feature representations (Table 1). Despite clinical evidence of equivalent survival outcomes, models trained solely using foundation model extracted features showed differences in AUC, and accuracy, accompanied by high standard deviation. Such regimen-specific instability suggests these models capture irrelevant features. PRISM's minimal accuracy fluctuation (Δ = 1.44%) demonstrates alignment with the trial's finding of no survival benefit differences between treatment regimens.

The Kaplan-Meier survival curves provide compelling visual evidence of PRISM's superior prognostic stratification capabilities, demonstrating the most pronounced and consistent separation between high-risk and low-risk survival curves throughout the five-year follow-up period with minimal convergence, which directly correlates with our quantitative metrics (Figure 3).

The sustained separation observed in PRISM's curves has important clinical implications, as the early differences of survival probabilities within the first year post-resection suggests that morphology-informed features can identify high-risk patients who may benefit from more aggressive adjuvant therapy or closer surveillance protocols, while the maintained separation throughout the five-year follow-up period indicates that these morphological patterns capture fundamental tumor biology characteristics that influence long-term outcomes rather than short-term artifacts. The superior survival curve separation achieved by PRISM can be attributed to its integration of CRC-specific morphological features with histopathological patterns, enabling quantification of tumor morphology and invasiveness, tumor microenvironment characteristics, and stromal patterns that capture biologically relevant information translating into meaningful prognostic differences. The smooth, consistent nature of PRISM's survival curves, without the erratic fluctuations observed in benchmark methods, indicates robust feature extraction and stable risk prediction crucial for clinical implementation, suggesting that the model's prognostic assessments are less susceptible to noise or artifacts in histopathological images and supporting its potential for reliable clinical decision-making in diverse patient populations. These findings collectively indicate that effective clinical implementation requires models that simultaneously optimize for: (1) balanced sensitivity/specificity, (2) cross-validation stability, and (3) biologically relevant feature learning.

## 3.2. Our New Standards for Model Validation and Bias Mitigation

The balanced sensitivity (67.14% ± 8.26%) and specificity (68.86% ± 8.62%) achieved by PRISM addresses a critical challenge in clinical implementation, where extreme performance on one metric at the expense of the other can limit practical utility. Unlike benchmark methods that exhibited 10-15% differences between these metrics, our AI prognostic model provides reliable predictions across both outcomes in correctly identifying patients at high risk of death within five years while minimizing false alarms that could lead to overtreatment.

The robustness of PRISM across diverse clinical subgroups strengthens its translational potential. The minimal performance variation across sex subgroups (AUC difference of only 0.03 between male and female patients; Supplementary Results: 7.2) and consistent performance across histological grades demonstrate that the model captures fundamental biological features rather than demographic artifacts. Similarly, reproducing the Alliance cohort clinical trial results, with only 1.44% accuracy difference between FL and IFL regimens, aligns with clinical trial evidence showing equivalent survival outcomes and suggests that our model learns treatment-independent morphological features.

Our analysis reveals a nuanced relationship between tumor location, sample size, and predictive performance that has important implications for clinical deployment (Supplementary Results: 7.1). The superior performance in sigmoid colon cases (n=149; AUC: 0.77 ± 0.06) compared to smaller anatomical subgroups suggests that both adequate sample representation and location-specific morphological patterns contribute to model reliability. The poor performance in hepatic flexure cases (n=28; AUC: 0.56 ± 0.30) likely reflects both limited training examples and potentially distinct biological characteristics of tumors arising in different anatomical locations. These findings highlight the need for location-stratified model development or the incorporation of anatomical location as an explicit feature in future iterations. From a practical standpoint, this suggests that initial clinical deployment might prioritize the most common tumor locations where adequate training data exists, with subsequent expansion as larger datasets become available for rarer anatomical sites.

Our study also exposes critical flaws in conventional validation approaches for survival prediction models in histopathology. The inadequacy of standard K-fold cross-validation for this application stems from its failure to account for the complex demographic and socioeconomic factors that can introduce subtle biases into histopathological data. The proposed stratified sampling approach based on demographic clustering represents a significant methodological advance that should be adopted more broadly in the field (Supplementary Results: 7.3). The discovery that AI models can inadvertently learn and amplify socioeconomic biases present in WSIs—factors often imperceptible to pathologists—represents a critical finding with broad implications for clinical AI development. These biases, linked to race, age, BMI, and income, can lead to disparate performance across patient populations, potentially exacerbating healthcare inequities. Our stratified K-fold validation approach provides a rigorous framework for identifying and mitigating these biases before clinical deployment.

### 3.3. Limitations and Challenges

Our study introduces an interpretable AI prognostic model that incorporates morphology-informed features for five-year survival prediction and addresses a critical gap in the interpretability of computational pathology models. PRISM's interpretability stems from its ability to capture the continuous variability spectrum within distinct morphological patterns. This enables clinicians to understand the biological basis underlying model predictions, as each extracted feature correlates with specific tissue morphologies that pathologists can visually validate.

The interpretable nature of our morphology-informed features provides insights into how patient subgroups differ at the time of surgical resection, potentially serving as prognostic biomarkers for survival prediction. These morphological signatures reflect underlying tumor biology, micro-environmental characteristics, and disease progression patterns that influence long-term outcomes. There are several limitations that constrain our ability to establish definitive causal relationships between resection-time morphological features and survival outcomes. This is because there is a temporal disconnect between patient surgical resection and treatment administration. Since surgical resection precedes extended adjuvant therapy protocols, the morphological features captured at resection may not fully account for treatment-related biological changes that occur over the subsequent months to years. This temporal gap introduces uncertainty regarding whether the identified morphological patterns represent features that may be modified by subsequent therapeutic interventions.

The relationship between training data scale and model performance in computational pathology presents a compelling question that deserves thoughtful examination through the lens of therapeutic dynamics. The observation that training data scale may become less consequential when models lack treatment awareness presents an intriguing consideration rooted in fundamental cancer biology. The temporal disconnect between surgical resection and subsequent different adjuvant therapy administration creates a critical biological gap that even massive datasets cannot bridge. During this interval, profound cellular and molecular changes occur: chemotherapy-induced DNA damage responses alter cellular morphology, stromal remodeling modifies the tumor microenvironment architecture, and selective pressure from cytotoxic agents drives clonal evolution and emergence of resistant subpopulations. These treatment-induced biological transformations include changes in tumor-infiltrating lymphocyte populations, vascular remodeling, fibroblast activation states, and epithelial-mesenchymal transition dynamics that fundamentally alter the morphological landscape that ultimately determines patient outcomes.

This raises meaningful questions about whether a single, treatment-naive foundation model, regardless of its training scale, may face fundamental constraints in prognostic capacity given the dynamic nature of cancer biology. Such models primarily capture static morphological patterns representing a single temporal snapshot of tumor heterogeneity, cellular architecture, and microenvironmental organization, while remaining blind to the evolving biological processes that drive therapeutic resistance, metastatic progression, and treatment response. The complex interplay between tumor cell intrinsic factors (oncogene addiction, DNA repair capacity, metabolic plasticity), extrinsic microenvironmental influences (immune cell infiltration, stromal composition, hypoxic gradients), and treatment-induced selective pressures creates a dynamic biological system that cannot be adequately represented by pre-treatment morphological features alone. This suggests that future developments might benefit from thoughtfully exploring multi-modal, temporally-aware models that integrate morphological patterns with molecular signatures (genomic alterations, transcriptomic profiles, proteomic landscapes), treatment response biomarkers (circulating tumor DNA dynamics, immune activation markers), and longitudinal biological data across the entire therapeutic continuum, thereby capturing the full spectrum of cancer biology rather than simply scaling the size of foundation models.

Several limitations warrant consideration. First, our study focuses exclusively on stage III CRC patients from a single clinical trial, which limits generalizability to other stages and practice settings. The Alliance cohort [35], while well-characterized, represents a selected population that met specific trial inclusion criteria that may differ from routine clinical practice populations. Second, the morphology-aware classifier was trained on the Hist-AI colorectal dataset, which, while comprehensive, may not capture all morphological variations encountered in diverse clinical settings. Future work should explore training on larger, more diverse morphological datasets to enhance feature extraction capabilities. Third, our analysis reveals performance limitations in less common anatomical subgroups in our patient group, suggesting the need for larger datasets or alternative modeling approaches for comprehensive coverage of all tumor locations. The development of meta-learning approaches that can effectively learn from limited data in rare anatomical locations represents an important future direction. Lastly, to establish robust interpretable biomarkers, future investigations require a more comprehensive data collection approach. This would necessitate molecular profiling at multiple timepoints throughout the treatment continuum, enabling researchers to distinguish between morphological features that remain stable predictors versus those that evolve with treatment response. Such holistic data collection strategies would strengthen the foundation for developing truly interpretable and clinically actionable prognostic biomarkers in CRC.

### 3.4. Broader Impact and Future Work

The integration of morphology-aware, high-level AI features with conventional histopathological characteristics represents a paradigm shift toward more sophisticated, biologically informed AI systems in oncology. Our AI prognostic model moves beyond the current limitations of "black box" deep learning models by incorporating domain-specific knowledge in a principled manner. The success of our AI prognostic model suggests broader applications across other cancer types where morphological heterogeneity plays a critical role in prognosis. The rigorous bias detection and mitigation strategies developed in this study provide a template for responsible AI development in healthcare. As AI systems become increasingly integrated into clinical workflows, the proactive identification and correction of algorithmic biases will be essential for ensuring equitable healthcare delivery.

Overall, PRISM presents a significant advancement in AI-driven CRC prognostication, demonstrating that morphology-aware, high-level feature extraction can substantially improve survival prediction accuracy. Beyond its technical contributions, this study establishes new standards for model validation and bias mitigation in clinical AI applications. PRISM's robust performance across diverse patient subgroups, combined with its principled approach to addressing algorithmic bias, positions it as a promising tool for clinical translation. Future work will focus on extension to other cancer stages and types, and continued refinement of bias mitigation strategies to ensure equitable deployment across diverse patient populations.

## 4. Datasets and Preprocessing
### 4.1. Ethics Statement and Patient Cohorts

For our main analysis, we used data from Cancer and Leukemia Group B (CALGB) 89803, a randomized phase III trial that compared adjuvant weekly bolus IFL versus FL alone in 1,264 patients with completely resected stage III CRC (Identifier: NCT00003835; Registry Identifier: NCI-2012-01844; First Submitted: November 1, 1999; First Posted: April 27, 2004; Study Start: May 1999). CALGB is now part of the Alliance for Clinical Trials in Oncology, so this is referred to as the Alliance cohort. Eligible patients had no prior chemotherapy/radiotherapy history, performance status 0–2, and initiated treatment within 21–56 days post-resection. Stratification factors included lymph node involvement (1–3 vs ≥4 nodes), histology grade, and preoperative CEA. The FL arm received the Roswell Park regimen (Leucovorin (LV) 500 mg/m² + 5-Fluorouracil (5FU) 500 mg/m² weekly × 6, every 8 weeks for 4 cycles). The IFL arm received Irinotecan (CPT-11) 125 mg/m² + LV 20 mg/m² + 5FU 500 mg/m² weekly × 4, every 6 weeks for 5 cycles. Primary endpoints were overall survival (OS) and disease-free survival (DFS). The trial was approved by institutional review boards at all participating centers in cooperative groups CALGB, NCCTG, NCIC CTG, ECOG, SWOG, and all patients provided written informed consent. We have provided the detailed characteristics of the dataset in Table 2.

**Table 2.** Demographic characteristics of Alliance Cohort

## Alliance Cohort Characteristics

| | | |
|---|---|---|
| Number of slides | | 431 |
| Number of Patients | | 424 |
| Mean age (yrs) | | 60.47 |
| Mean of household income median (USD) | | 43194.59 |
| Race | White | 398 |
| | Black | 33 |
| Sex | Male | 240 |
| | Female | 191 |
| Treatment | 5FU/LV | 219 |
| | CPT-11/5FU/LV | 212 |
| Zubrod Performance scale | 0 | 328 |
| | 1 | 98 |
| | 2 | 2 |
| Tumor location | Cecum | 101 |
| | Ascending Colon | 64 |
| | Hepatic Flexure | 28 |
| | Transverse Colon | 46 |
| | Splenic flexure | 19 |
| | Descending colon | 19 |
| | Sigmoid colon | 149 |
| Grade | I | 20 |
| | II | 300 |
| | III | 108 |
| | IV | 0 |
| Stage | I | 6 |
| | II | 42 |
| | III | 350 |
| | IV | 8 |
| | V | 20 |
| Small blood/lymphatic vessel invasion | No | 285 |
| | Yes | 138 |
| Extramural vascular invasion | No | 387 |
| | Yes | 29 |
| Infiltrating border | No | 274 |
| | Yes | 141 |
| Five-year survival | Yes | 29 |
| | Survived | 328 |

### 4.2. Tissue Scanning, and Quality Assurance Protocol

All tissue samples in the Alliance cohort were resected from one of the following tumor locations: cecum, ascending colon, hepatic flexure, transverse colon, splenic flexure, descending colon, or sigmoid colon. Following resection, all samples underwent formalin fixation and were subsequently stained with H&E. WSIs of Alliance cohort were then acquired through digital scanning at 40× magnification (0.25 μm/pixel resolution) using an Aperio Digital Pathology Scanning system (Leica Biosystems) at The Ohio State University, Wexner Medical Center, Columbus, Ohio. This study received approval from the Ohio State University Institutional Review Board (IRB 2018C0098) with a waiver for informed consent. These WSIs were individually, manually reviewed by a trained expert pathologist to ensure that a tumor was present. In over >98% cases had one representative tumor slide/WSI for evaluation. Exclusion criteria include: WSI with no tumor, lymph node tissue only, or mucinous (tumor composed of pools of mucin with floating tumor cells only) cases, and those with death attributed to causes other than disease from within five-years deceased group. Cases were also reviewed for quality check to ensure correct tissue detection and coverage, color fidelity, focus/sharpness, and no scanning artifacts; WSIs with inadequate staining were also removed. This comprehensive quality control process resulted in the exclusion of 211 cases from the original Alliance cohort, leaving 424 cases for PRISM training and evaluation. Throughout the entire scanning and tissue quality assessment process, reviewing pathologists remained blinded to all patient clinical information and outcomes. This rigorous filtering ensures that the dataset used for training PRISM is high-quality, relevant, and consistent, which is crucial for reliable prognostication.

## 5. Methods

### 5.1. Identification of Subgroups with Varied Survival Rates and Stratified Data Splitting for Effective Generalization

A critical aspect of developing robust prognostic models involves understanding the heterogeneity within patient populations and ensuring that model training and evaluation account for clinically relevant subgroups. Traditional approaches to model validation often overlook the inherent diversity in patient demographics and clinical characteristics, which can lead to biased performance estimates and reduced generalizability across different patient populations. To address this fundamental limitation, we conducted a comprehensive analysis of patient subgroups within our Alliance cohort to identify distinct populations with varying survival outcomes. We then implemented a stratified approach to data splitting that preserves the representation of these diverse patient characteristics.

Our quantitative analysis of clinical data revealed distinct patient subgroups with different survival outcomes across multiple demographic and clinical parameters. We systematically grouped patients based on several key criteria and assessed five-year survival rates within each subgroup. Age stratification demonstrated significant prognostic value, with patients aged ≤65 years exhibiting substantially different five-year survival rates compared to those >65 years. This age-based survival disparity reflects the complex interplay between chronological age, comorbidity burden, treatment tolerance, and overall physiological reserve in CRC patients. Similarly, we identified significant survival variations based on socioeconomic factors, particularly income levels, which serve as a proxy for healthcare access, treatment compliance, and overall health status. The income-based stratification revealed distinct survival patterns that likely reflect disparities in healthcare quality, early detection rates, and access to optimal treatment regimens. Body mass index (BMI) stratification also demonstrated prognostic significance, with patients categorized into three distinct groups: BMI <25 kg/m², BMI 25-30 kg/m², and BMI ≥30 kg/m². These BMI-based subgroups exhibited different five-year survival rates, potentially reflecting the complex relationship between nutritional status, metabolic health, treatment toxicity, and surgical outcomes in CRC patients.

To ensure robust model development and evaluation, we implemented a stratified data splitting approach that maintained proportional representation of these clinically relevant subgroups across training and validation sets (figure 5). As a result, PRISM prevents the inadvertent creation of training sets that over-represent or under-represent specific patient populations, thereby ensuring that PRISM's prognostic performance is evaluated against the full spectrum of patient diversity present in clinical practice. This approach addresses the critical limitation of conventional validation strategies that may produce overly optimistic performance estimates by failing to account for population heterogeneity and potential subgroup-specific biases embedded within histopathological images.

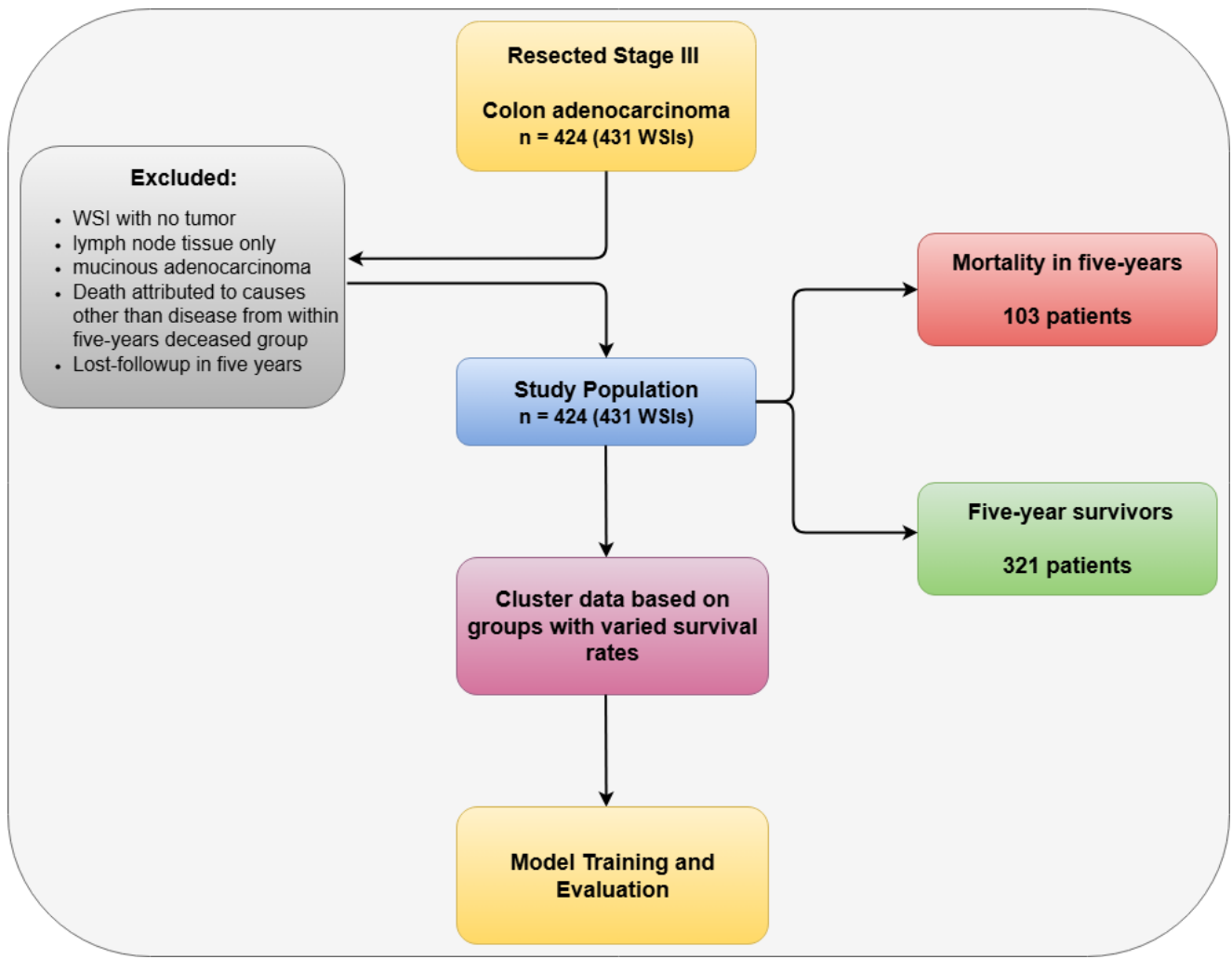

**Figure 5: Study population and cohort design.** A Flowchart illustrating the patient selection process for colon carcinoma cases, including exclusion criteria: whole-slide images (WSIs) with no tumor, lymph node tissue only, mucinous adenocarcinoma, non-disease-related deaths within five years, and cases lost to follow-up within five years. The final cohort included 424 patients with 431 WSIs, stratified into deceased (n=103) and survivor (n=321) groups based on five-year follow-up. To ensure balanced representation across varying survival rate groups, patients were clustered based on age, BMI, and income, and five-folds were constructed within each cluster. These were then concatenated to form training, validation, and test sets.

### 5.2. Morphology-Informed Classifier and Feature Extractor

Current foundation models in computational pathology have demonstrated strong generalization across tasks [2, 18], but they do not explicitly encode morphological semantics. These models typically rely on self-supervised learning to capture broad visual representations, often overlooking the domain-specific morphological features (biomarkers) that are critical for accurate prognostication. In contrast, pathologists do recognize these patterns but tend to describe them in discrete terms—labeling regions as "neoplastic," "stromal," or "inflammatory"—without accounting for the continuous phenotypic variation that exists within these categories. Existing computational systems that attempt to extract morphological features often follow a two-step process: first, they classify tissue regions into predefined categories; then, they compute basic statistics (e.g., area, density) within those identified regions. While useful, this approach treats morphology as static and compartmentalized, failing to capture the gradual transitions and phenotypic diversity that reflect tumor evolution and biological complexity.

PRISM addresses this critical gap by explicitly modeling the continuous morphological variability within and across histological regions. Rather than relying solely on categorical labels or generic feature extraction, PRISM learns to represent nuanced phenotypic differences—such as subtle changes in nuclear morphology, glandular architecture, and stromal composition—and integrates these with generic histopathological features to construct a more biologically faithful and prognostically powerful representation. To extract these morphology-informed features, we developed a deep learning model trained on the publicly available Histopathology AI (Hist-AI) colorectal dataset [25]. Hist-AI dataset contains 13 distinct tissue morphologies, including: High-Grade Adenocarcinoma, Low-Grade Adenocarcinoma, High-Grade Adenoma, Low-Grade Adenoma, Fat, Hyperplastic Polyp, Inflammation, Mucin, Smooth Muscle, Necrosis, Sessile Serrated Lesion, Stroma, and Vascular Structures. We used 224 × 224-pixel patches extracted at 20× magnification to train a morphology classification network.

During training, PRISM not only learns to classify these morphologies but also captures the intra-class variability within each category. This enables the model to extract a diverse set of high-dimensional features that reflect the phenotypic spectrum within each morphological type. The network was optimized using cross-entropy loss and evaluated using five-fold cross-validation to ensure generalization. For downstream prognostication, the final classification layer was removed, and the penultimate fully connected layer was used as a feature extraction backbone (Figure 6). These learned features—rich in tissue-specific morphological information—serve as the foundation for PRISM's survival prediction model, enabling it to capture clinically relevant patterns that are often missed by traditional approaches.

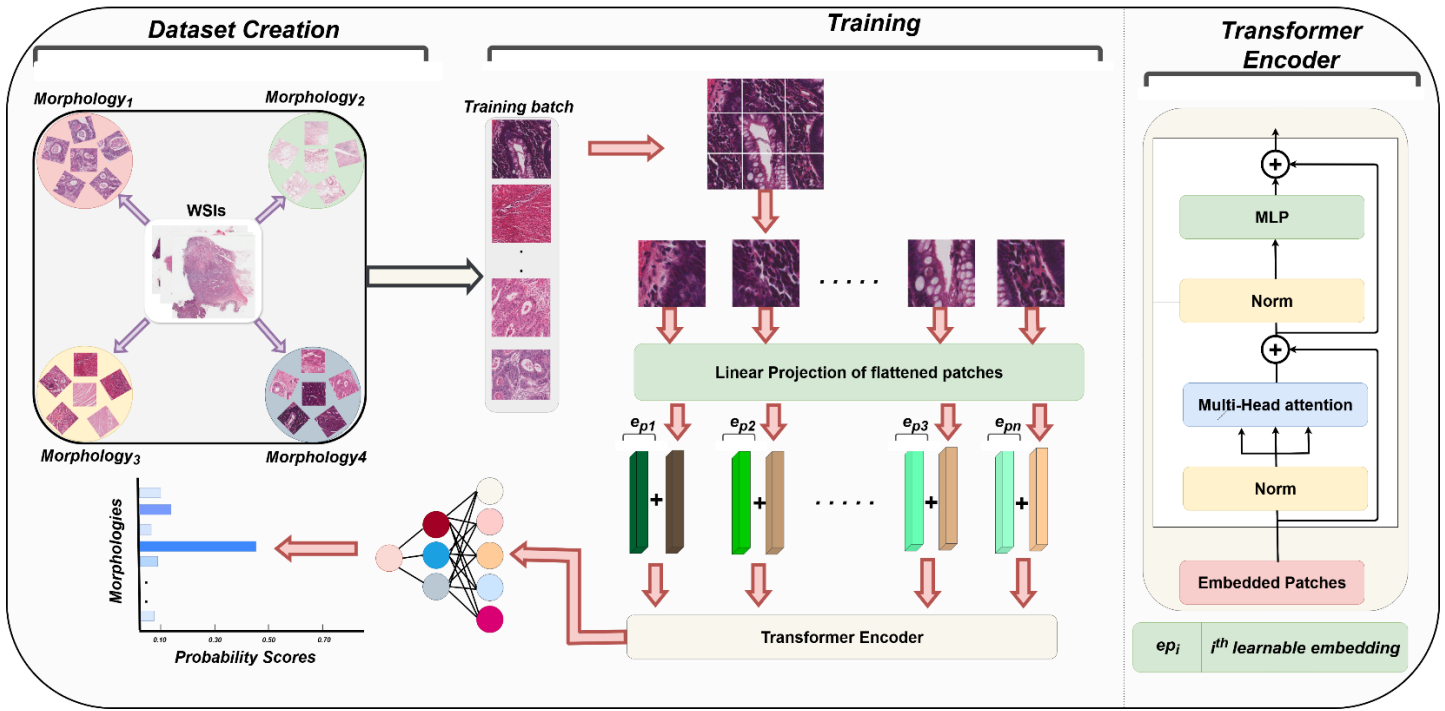

**Figure 6. Overview of PRISM's morphology classifier training pipeline.** Whole slide images (WSIs) are first annotated by expert pathologists to identify 13 distinct morphological tissue regions (Hist-AI dataset). Histological patches are then extracted from these annotated regions and used to train a morphology classification module. Once trained, this module generates specialized morphology-informed features, which PRISM integrates with generic histopathological features to construct a robust, multi-faceted representation for five-year survival prediction in stage III CRC.

### 5.3. Morphology Aware Survival Prediction

Building upon the morphology-informed feature extraction capabilities described above, PRISM integrates these domain-specific features with foundation model embeddings to construct a comprehensive AI prognostic model. Foundation models [2], trained on large-scale histological datasets using self-supervised learning, capture broad visual patterns but lack explicit encoding of biologically meaningful morphological semantics. In contrast, PRISM's morphology-informed features are derived from expert-annotated tissue regions and capture nuanced phenotypic variations—such as differences in nuclear morphology, glandular architecture, and stromal composition—that are critical for prognostication. By combining the generalizability of foundation models with the specificity of morphology-aware representations, PRISM creates a more biologically faithful and prognostically powerful feature space.

To effectively model survival outcomes, PRISM employs a multi-instance learning (MIL) paradigm [1], which accommodates the variable number of tissue patches per WSI and learns to weigh the relative importance of different morphological patterns. This approach addresses the critical need for prognostic models that can capture the complex interplay between diverse tissue morphologies while accounting for the inherent spatial and phenotypic heterogeneity present within CRC slides. The resulting model not only improves predictive performance but also enhances interpretability by linking prognostic signals to specific morphological contexts.

Here are the implementation details of PRISM. Let $X = \{X_1, X_2, .., X_n\}$ represent the $n$ patient WSIs with five-survival $Y = \{y_1, y_2, ..., y_n\}$, where $y_n \in \{0, 1\}$ for $i = \{1, ..., n\}$ corresponds to the five-survival label with 1 means that patient died within five years and 0 means that patient survived for five-years post resection. For each WSI $X_i$, for $i = \{1, ..., n\}$, we divide $X_i = \{x_{i1}, x_{i2}, .., x_{im}\}$ into $m$ patches with each patch $x_{i1}$ representing 224 × 224 area at 20× magnification, and patches with less than 25% of the tissue were excluded using TRIDENT [26, 27]. Then, a vision transformer-based histopathology trained feature extractor (UNI) [2, 28] is used to extract the features $g_{ij}$ from each patch $x_{ij} \in \mathbb{R}^d$ of the WSI $X_i$. Subsequently, we also feed each patch $x_{ij}$ of the WSI $X_i$ from morphology informed feature extractor and extract the morphology aware features $m_{ij}$ for each patch $x_{ij} \in \mathbb{R}^d$ of WSI.

Once we extract both features ($g_{ij}$ and $m_{ij}$) from each patch $x_{ij}$, we compute the cross-feature interaction of each morphology aware feature with each foundation model feature to generate $f'_{ij}$ where $f'_{ij} \in \mathbb{R}^{d \times d}$ (figure 1). Then, we use a neural network to project them again as $f_{ij} \in \mathbb{R}^d$ to merge the morphology informed features with foundation model features. Specifically, we learn a transformation T: $\mathbb{R}^{d \times d} \to \mathbb{R}^d$, that maximizes the mutual information $I(f_{ij}; f'_{ij}) = I(f_{ij}; T(f_{ij}))$ while enforcing the constraint $d \times d' \ll d$. Since each slide $X_i$ contains a different number of patches, and each patch may have varying prognostic significance, we employ multiple instance learning (MIL) aggregation [1] to combine the morphology-informed patch features and generate the slide-level representation $Z_i$ (eq. 1),

$$Z_i = \sum_{j=1}^{n} a_{ij} f_{ij} \quad (1)$$

where

$$a_{ik} = \frac{exp(W^T(\tanh(V^T f_{ik})) \odot V^T(\text{sigm}(U^T f_{ik})))}{\sum_{j=1}^{n} exp(W^T(\tanh(V^T f_{ij})) \odot V^T(\text{sigm}(U^T f_{ij})))} \quad (2)$$

is the importance score $a_{ik}$ computed for each patch $x_{ik}$, which quantifies the relative prognostic relevance of patch $x_{ik}$ with respect to other patches within the WSI $X_i$ (Eq. 2). Here, $V \in \mathbb{R}^{d \times l}$, $U \in \mathbb{R}^{d \times l}$, and $W \in \mathbb{R}^{l \times 1}$ are the learnable parameters of the attention network, $l$ represents the number of neurons in the hidden layer, and $\odot$ denotes element-wise multiplication. This morphology-informed slide representation $Z_i$ is subsequently used by a separate neural network to predict five-year survival. During the training process, all cross-modal interactions between generic histopathological features and morphological informed features for each patch $x_{ij}$ are learned and optimized in an end-to-end fashion, which enables optimized of feature combinations most relevant for prognostic prediction.

## 5.4. Evaluation Protocols and Implementation Details

To train our morphology-informed feature extraction module, we utilized the publicly available Hist AI colorectal dataset [25]. This dataset comprises 77,182 annotated histological patches derived from 1,719 H&E-stained WSIs, systematically categorized into 13 distinct morphological classes. For training, we used 70% of the histological patches from each morphological class, with 15% allocated for validation and the remaining 15% for testing. This stratified sampling approach ensured balanced representation across all tissue types in each data split.

For training PRISM, we partitioned each WSI into 224 × 224 × 3 patches at 20× magnification using TRIDENT [26, 27] and excluded patches containing less than 25% tissue content. Then, we extracted generic histopathological features using a foundation model (UNI) [2] to ensure consistent feature representation across all comparative approaches [1, 29-32]. For training the morphology-informed classifier, we implemented the Hibou foundation model [21] as a feature extractor, coupled with a multi-layer perceptron (MLP) head comprising two fully connected layers of 512, 128, and 13 dimensions, respectively, utilizing ReLU and SoftMax activation functions. Once morphology-informed classifier was trained, we obtained morphology-informed features by truncating the model after the 512-dimensional layer to capture hidden feature representations. Subsequently, we configured PRISM training with the following hyperparameters: learning rate of $2\times10^{-5}$, Adam optimizer [33], Xavier uniform weight initialization [34], batch size of 1, and L1 regularization coefficient of $5\times10^{-4}$. For comparative evaluation, we implemented baseline methods using default parameters from their respective GitHub repositories, with the exception of RRT-MIL [31], where we computed classification thresholds on the validation set to optimize performance evaluation. We omitted dropout regularization as our model incorporated L1 regularization to promote feature sparsity and prevent overfitting to irrelevant morphological patterns. We employed Xavier uniform initialization [34] to ensure appropriate gradient flow and stable training dynamics throughout the deep architecture and maintained training reproducibility through fixed random seed implementation, enabling consistent results across multiple experimental iterations.

We evaluated the performance of PRISM on: (i). Alliance cohort; (ii). a publicly available dataset using The Cancer Genome Atlas (TCGA) CRC cohort, a multicenter study encompassing patients with stage I–IV disease, predominantly from institutions across the United States. All histopathological images and associated clinical data from the TCGA study are publicly accessible through the Genomic Data Commons portal (https://portal.gdc.cancer.gov)."

For each cohort, we used 70% patients for training, 10% patients for validation and 20% for testing using stratified K-fold cross validation within each cluster. Specifically, we trained our model separately on each cohort using cohort specific training data and morphology-informed feature extractor to evaluate the performance. To compare the statistical significance between methods, a two-sided Wilcoxon signed-rank test was used.

**Clinicaltrials.gov Identifier**: NCT00003835 (CALGB 89803); **Registry:** CTRP (Clinical Trial Reporting Program); **Registry Identifier:** NCI-2012-01844

**Support:** The data from CALGB 89803 were obtained directly from the Alliance for Clinical Trials in Oncology, a National Clinical Trials Network cooperative group; however, all analyses and conclusions in this manuscript are the sole responsibility of the authors and do not necessarily reflect the opinions or views of the clinical trial investigators, the NCTN, the NCORP or the NCI.

## Author contributions

U.S. performed data preprocessing, experimental design, validation, and manuscript writing. A.R. and Z.S. provided feedback on study design, validation, and manuscript editing. W.F., W.C., and D.K. contributed clinical assessment, while W.F, and M.N.G. assisted with manuscript editing. W.F., A.R., W.C., D.K., U.S., M.K.K., and M.N.G. performed result analysis. W.C. and M.K.K.N. conceptualized and designed the study, oversaw validation, supervised the research, and contributed to manuscript editing.

## Acknowledgments

The authors gratefully acknowledge the Ohio Supercomputer Center for providing high-performance computing resources under its contract with The Ohio State University College of Medicine. We also thank the Department of Pathology and the Comprehensive Cancer Center at The Ohio State University for their valuable support.

## Funding

The project described was supported in part by R01 CA276301 (PIs: Niazi and Chen) from the National Cancer Institute, The project was also supported partly by Pelotonia under IRP CC13702 (PIs: Niazi, Vilgelm, and Roy), The Ohio State University Department of Pathology and Comprehensive Cancer Center. The content is solely the responsibility of the authors and does not necessarily represent the official views of the National Cancer Institute or National Institutes of Health or The Ohio State University.

## Competing Interests

All authors declare no financial or non-financial competing interests.

## Code Availability

The underlying code for this study is available in AI4Path PRISM repository and can be accessed via following link: https://github.com/AI4Path-Lab/PRISM.

## Ethics Approval and Consent to Participate

This study was reviewed and approved by the Institutional Review Board of Ohio State University (IRB #2018C0098). All procedures involving human data were conducted in accordance with the ethical standards of the institutional research committee, the national research regulations, and with the 1964 Declaration of Helsinki and its later amendments. Given the retrospective nature of the study and because data were archival and de-identified, the requirement for informed consent to participate was formally waived by the Institutional Review Board.

# Supplementary Materials

# 7. Additional Results:

## 7.1. Tumor Location-Dependent Survival Prediction Using PRISM: Larger Patient Numbers Offset Batch Effects and Enhance Performance

To assess the generalizability and clinical applicability of PRISM across different colorectal anatomical sites, we conducted comprehensive subgroup analysis stratified by tumor resection location. This evaluation is critical for understanding how morphological feature patterns and model performance vary across the diverse biological microenvironments within the colon, informing clinical deployment strategies and identifying potential limitations in site-specific prognostication.

The prognostic performance across colon tumor resection cohorts exhibited a strong correlation with total sample size (n), where larger patient groups stratified by tumor location yield more reliable prognostic performance, though inherent biological differences in feature patterns across anatomical locations may also contribute to performance variations. As detailed in Table S1 and Figure S1, in patients with sigmoid colon cancers (n=149), PRISM achieved optimal performance (AUC: $0.77 \pm 0.06$; C-index: 0.75; HR: 3.88). Following this, in cecal cancers (n=101), PRISM demonstrated moderate but stable metrics (AUC: $0.71 \pm 0.11$; C-index: 0.69; HR: 2.97), showing the performance drop despite adequate sampling. This performance decline accelerated with ascending colon cancers (n=64), where results remained robust (AUC: $0.74 \pm 0.21$; C-index: 0.71; HR: 3.10) but exhibited high standard deviation. In transverse colon cancers (n=46), PRISM demonstrated high sensitivity ($84.52 \pm 15.5$) but critically low specificity ($50.01 \pm 17.46$), suggesting location-specific feature importance may skew predictions. Conversely, smaller cohorts revealed significant limitations: splenic flexure cancers (n=19) had the poorest AUC ($0.42 \pm 0.43$) and C-index (0.49), while hepatic flexure (n=28) and descending colon cancers (n=19) showed near-random sensitivity ($50.0 \pm 50.0$) and marginal hazard ratios (3.29 and 2.39) with large 95% confidence intervals (Figure S1), indicating that both limited samples and possibly local specific features may compromise model generalizability. These results collectively demonstrate that reliability of deep learning models depends critically on cohort size, and tumor microenvironment differences across locations, necessitating location-specific feature extraction for clinically viable models.

**Table S1:** Five-year OS results of PRISM stratified by tumor location in the Alliance cohort using five-fold cross-validation. Each row reports the average performance with standard deviation. For each metric, the best result is shown in bold, and the second best is underlined.

|  | Tumor location | AUC | Accuracy (%) | Sensitivity (%) | Specificity (%) | No of samples (n) |
|---|---|---|---|---|---|---|
| **Right Colon** | Cecum | $0.71 \pm 0.11$ | $67.58 \pm 6.70$ | $67.85 \pm 0.20$ | $67.32 \pm 11.00$ | 101 |
|  | Ascending Colon | $0.74 \pm 0.21$ | **$72.46 \pm 20.32$** | $74.28 \pm 0.24$ | $70.64 \pm 17.39$ | 64 |
|  | Hepatic Flexure | $0.56 \pm 0.30$ | $60.77 \pm 22.22$ | $60.95 \pm 25.71$ | $60.60 \pm 22.81$ | 28 |
|  | Transverse Colon | $0.74 \pm 0.19$ | $67.26 \pm 3.80$ | **$84.52 \pm 15.5$** | $50.01 \pm 17.46$ | 46 |
|  | Right Colon | $0.69 \pm 0.05$ | $68.08 \pm 3.59$ | **$71.62 \pm 10.12$** | $64.53 \pm 10.56$ | 239 |
| **Left Colon** | Splenic Flexure | $0.42 \pm 0.43$ | $65.83 \pm 15.87$ | $50.00 \pm 50.00$ | $81.66 \pm 18.48$ | 19 |
|  | Descending Colon | $0.46 \pm 0.27$ | $64.58 \pm 14.87$ | $50.00 \pm 50.00$ | **$79.16 \pm 21.65$** | 19 |
|  | Sigmoid Colon | **$0.77 \pm 0.06$** | $70.09 \pm 3.70$ | $64.69 \pm 8.40$ | $75.49 \pm 12.29$ | **149** |
|  | Left Colon | $0.71 \pm 0.08$ | $67.69 \pm 5.69$ | **$61.39 \pm 12.66$** | $73.99 \pm 10.03$ | 187 |

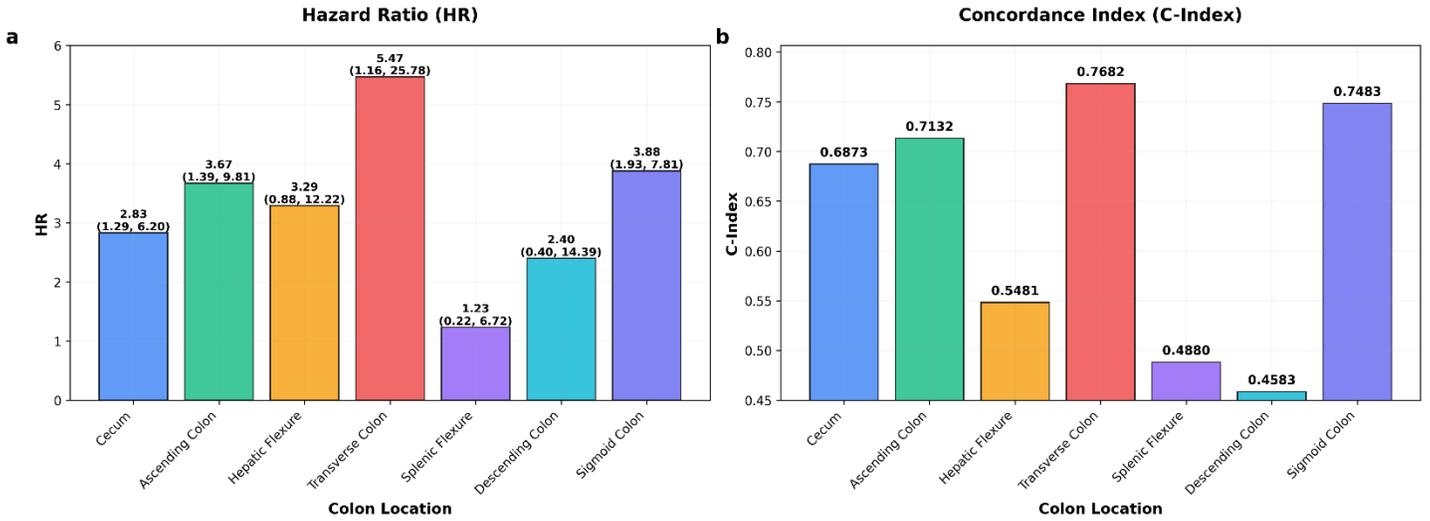

**Figure S1:** Subgroup analysis of our model's performance stratified by colon tumor location in the Alliance cohort using five-fold cross-validation. (a) Hazard ratios with 95% confidence intervals for each anatomical site, and (b) Concordance Index (C-Index) values demonstrating model's ability to risk-stratify patients across different colon segments. PRISM achieved the highest C-Index (0.7483) and robust hazard ratio (3.88) in sigmoid colon cancers, while smaller cohorts from splenic flexure and descending colon showed reduced performance, indicating that model reliability correlates with sample size and possibly anatomical site-specific biological features.

## 7.2. Robustness of PRISM Across Demographics and Histology

A critical requirement for clinical deployment of AI-based prognostic models is demonstrated robustness across diverse populations and clinical contexts. To address concerns about generalization and bias, we conducted comprehensive subgroup analyses of PRISM within the Alliance cohort, evaluating performance consistency across demographic and pathological stratifications. These analyses confirm PRISM's superior stability and reliability compared to benchmark methods (ABMIL, CLAM, Nakanishi et al., RRT-MIL).

PRISM exhibited strong consistency across sexes (Table S2), a key indicator of clinical robustness. Performance metrics remained stable between male (AUC: $0.69 \pm 0.06$; accuracy: $68.06\% \pm 5.00$) and female (AUC: $0.71 \pm 0.08$; accuracy: $67.91\% \pm 3.20$) cohorts, with minimal AUC variation ($\Delta = 0.02$) and negligible accuracy fluctuation (0.15% difference). In contrast, benchmarks showed significant sex-based instability: CLAM had a 5.5% accuracy drop between sexes, while Nakanishi et al. and RRT-MIL exhibited 1.5–2% accuracy reductions and extreme sensitivity variability (e.g., RRT-MIL sensitivity SD: ±30.91). This consistency underscores PRISM's resilience to sex-based distribution shifts, suggesting reliance on generalizable biological features rather than spurious correlations.

Notably, PRISM achieved superior performance in female patients (highest AUC: $0.71 \pm 0.078$; highest accuracy: $67.91\% \pm 3.20$), surpassing CLAM by 15% in accuracy while delivering better-balanced sensitivity/specificity ($64.00\% \pm 11.0 / 71.81\% \pm 10.00$ vs. $20.33\% \pm 12.13 / 85.16\% \pm 13.61$) (Table S2). Its higher sensitivity is clinically critical for identifying high-risk female patients requiring aggressive intervention. In male patients, PRISM outperformed benchmarks in practical utility: while TransMIL had a marginally higher AUC ($0.67 \pm 0.06$), its AUC is influenced by predicting majority of the samples as five-year survivors, whereas PRISM achieved ~8% higher accuracy ($68.06\% \pm 5.00$) with balanced sensitivity/specificity ($69.7\% \pm 9.00 / 66.58\% \pm 11.00$) and lower metric variability, avoiding trade-offs observed in competitors.

**Table S2:** Five-year survival prediction results of PRISM stratified by sex in the Alliance cohort using five-fold cross-validation. Each row reports the average performance with standard deviation. For each metric, the best result is shown in bold, and the second-best is underlined.

| Model | Sex | AUC | Accuracy (%) | Sensitivity (%) | Specificity (%) |
| --- | --- | --- | --- | --- | --- |

| Method | Sex | AUC | Accuracy | Sensitivity | Specificity |
|---|---|---|---|---|---|
| ABMIL [29] | Female | 0.63 ± 0.11 | 57.86 ± 10.13 | 51.48 ± 17.68 | 64.25 ± 07.50 |
| | Male | 0.61 ± 0.05 | 57.68 ± 05.75 | 46.04 ± 12.87 | 69.32 ± 07.46 |
| CLAM [1] | Female | 0.61 ± 0.16 | 52.75 ± 09.29 | 20.33 ± 12.13 | 85.16 ± 13.61 |
| | Male | 0.63 ± 0.03 | 58.35 ± 03.95 | 31.38 ± 08.05 | 85.33 ± 05.95 |
| TransMIL [30] | Female | 0.69 ± 0.12 | 54.18 ± 05.99 | 08.37 ± 11.99 | 01.00 ± 00.00 |
| | Male | 0.67 ± 0.06 | 52.67 ± 06.19 | 06.00 ± 12.00 | 99.35 ± 01.20 |
| Nakanishi et. Al [32] | Female | 0.60 ± 0.11 | 58.32 ± 09.70 | 53.43 ± 20.25 | 63.21 ± 07.87 |
| | Male | 0.61 ± 0.06 | 59.99 ± 05.78 | 56.01 ± 12.59 | 63.97 ± 03.65 |
| RRT-MIL [31] | Female | 0.56 ± 0.07 | 54.11 ± 05.35 | 45.98 ± 30.91 | 62.24 ± 24.09 |
| | Male | 0.59 ± 0.08 | 56.17 ± 06.29 | 44.19 ± 21.72 | 68.16 ± 25.79 |
| PRISM | Female | 0.71 ± 0.07 | 67.91 ± 03.20 | 64.00 ± 11.00 | 71.81 ± 10.00 |
| | Male | 0.69 ± 0.06 | 68.06 ± 05.00 | 69.70 ± 09.00 | 66.38 ± 11.00 |

PRISM also generalized robustly across histologic heterogeneity (Table S3). Despite known inter-observer variability in grading, performance remained consistent across glandular differentiation grades:

- **Grade I** (well differentiated, n=20): High accuracy (81.25% ± 6.25) and AUC (0.75 ± 0.25), though variability reflected sample limitations.
- **Grade II** (moderately differentiated, n=297): Stable accuracy (67.73% ± 3.31) and AUC (0.71 ± 0.03) with low standard deviation.
- **Grade III** (poorly differentiated, n=104): Robust accuracy (67.25 ± 4.72) and AUC (0.70 ± 0.10)

This consistency across biologically distinct grades—particularly the stability in Grade II (the largest subgroup)—demonstrates PRISM's capacity to learn generalizable pathologic features. PRISM establishes a new standard for robust prognostication, consistently matching or outperforming benchmarks across genders and histologic grades while avoiding sensitivity-specificity trade-offs. Its minimal performance fluctuation in key subgroups (ΔAUC <0.03 across sexes; <4% accuracy variability in Grade II and III), superior accuracy in males and leading female-cohort performance in comparison with comparison methods, and adaptability to histologic heterogeneity directly address clinical concerns about generalization and bias. These attributes position PRISM as a uniquely reliable model for clinical translation in CRC prognosis.

**Table S3:** Five-year survival prediction results of PRISM stratified by Grade in the Alliance cohort using five-fold cross-validation. Each row reports the average performance with standard deviation. For each metric, the best result is shown in bold, and the second-best is underlined.

| Grade | AUC | Accuracy (%) | Sensitivity (%) | Specificity (%) | No of patients (n) |
|---|---|---|---|---|---|
| I | 0.75 ± 0.25 | 81.25 ± 6.25 | 75.00 ± 25.00 | 87.50 ± 12.5 | 20 |
| II | 0.71 ± 0.03 | 67.73 ± 3.31 | 65.11 ± 6.92 | 70.35 ± 8.99 | 297 |
| III | 0.70 ± 0.10 | 67.25 ± 4.72 | 72.85 ± 18.90 | 61.64 ± 14.63 | 104 |

## 7.3. Enhanced Performance of Baseline Methods using Add-on Module Integration and PRISM Superiority

To rigorously evaluate the impact of our proposed patient heterogeneity module, we applied this add-on enhancement to existing state-of-the-art methods and compared their performance against PRISM. The integration of our module consistently improved the prognostic accuracy of baseline models across multiple clinical subgroups, though PRISM maintained superior performance in all comparisons.

When applied to CLAM, our add-on module substantially improved its five-year survival prediction accuracy by 7% compared to baseline, ensuring that model training incorporates clinically relevant subgroups (Supplementary Figure S2). In the FL treatment subgroup, CLAM's accuracy increased from $50.04 \pm 5.96\%$ to $63.29 \pm 4.45\%$, representing a 13.25% absolute improvement and reduced variability (Supplementary Table S4). Similarly, in female patients, CLAM's accuracy improved from $52.75 \pm 9.29\%$ to $65.10 \pm 8.70\%$ (a 12.35% gain) (Supplementary Table S5). However, these enhancements did not surpass PRISM's performance, which achieved an accuracy of $68.21 \pm 3.40\%$ in the FL subgroup and $67.91 \pm 3.20\%$ in females, representing a 4.92% and 2.81% improvement, respectively. ABMIL also demonstrated notable gains with our module. For FL-treated patients, accuracy improved from $62.02 \pm 13.76\%$ to $63.86 \pm 6.30$, a 1.84% increase with significantly reduced standard deviation (from 13.76% to 6.30%). Sex-stratified analysis also revealed an 8% accuracy improvement in female patients (from $57.86 \pm 10.13\%$ to $61.71\% \pm 8.20$). Despite these improvements, PRISM still outperformed enhanced ABMIL by 4.35% in the FL subgroup and 8.77% in the IFL subgroup. Nakanishi et al.'s method showed more modest gains, with FL treatment accuracy improving from $61.12 \pm 12.85$ to $62.21 \pm 4.70$ and IFL accuracy increasing from $56.02 \pm 5.29$ to $58.83 \pm 8.60$. RRT-MIL exhibited variable performance: FL accuracy improved from $55.42 \pm 8.54$ to $59.51 \pm 11.11$ (a 4.09% gain), but IFL accuracy decreased from $55.62 \pm 3.61$ to $51.18 \pm 9.42$ (a 4.44% drop), indicating instability.

The HR analysis further validated PRISM's dominance, achieving a robust HR of 3.34 (95% CI: 2.28–4.90)—1.5× greater risk discrimination than the state-of-the-art methods (Supplementary Figure S3). While CLAM improved marginally (HR: $1.96 \rightarrow 2.25$) but failed to approach PRISM's precision, as their underlying architectures could not fully leverage the continuous morphological spectrum essential for robust risk stratification. PRISM's tight confidence interval (spanning 2.62 vs. TransMIL's 9.98) enabled clinically actionable identification of high-risk patients with 3.34× greater mortality likelihood within five years.

Critically, the add-on module reduced performance variability across all baselines, as evidenced by lower standard deviations in key subgroups (Supplementary Table S4). For instance: CLAM's standard deviation in the FL subgroup decreased from ±5.96% to ±4.45%. ABMIL's FL subgroup variability dropped sharply from ±13.76 to ± 6.30%. Nakanishi et al.'s FL subgroup variability reduced from ±12.85 to ± 4.70%. This confirms that accounting for population heterogeneity mitigates algorithmic bias and enhances model stability.

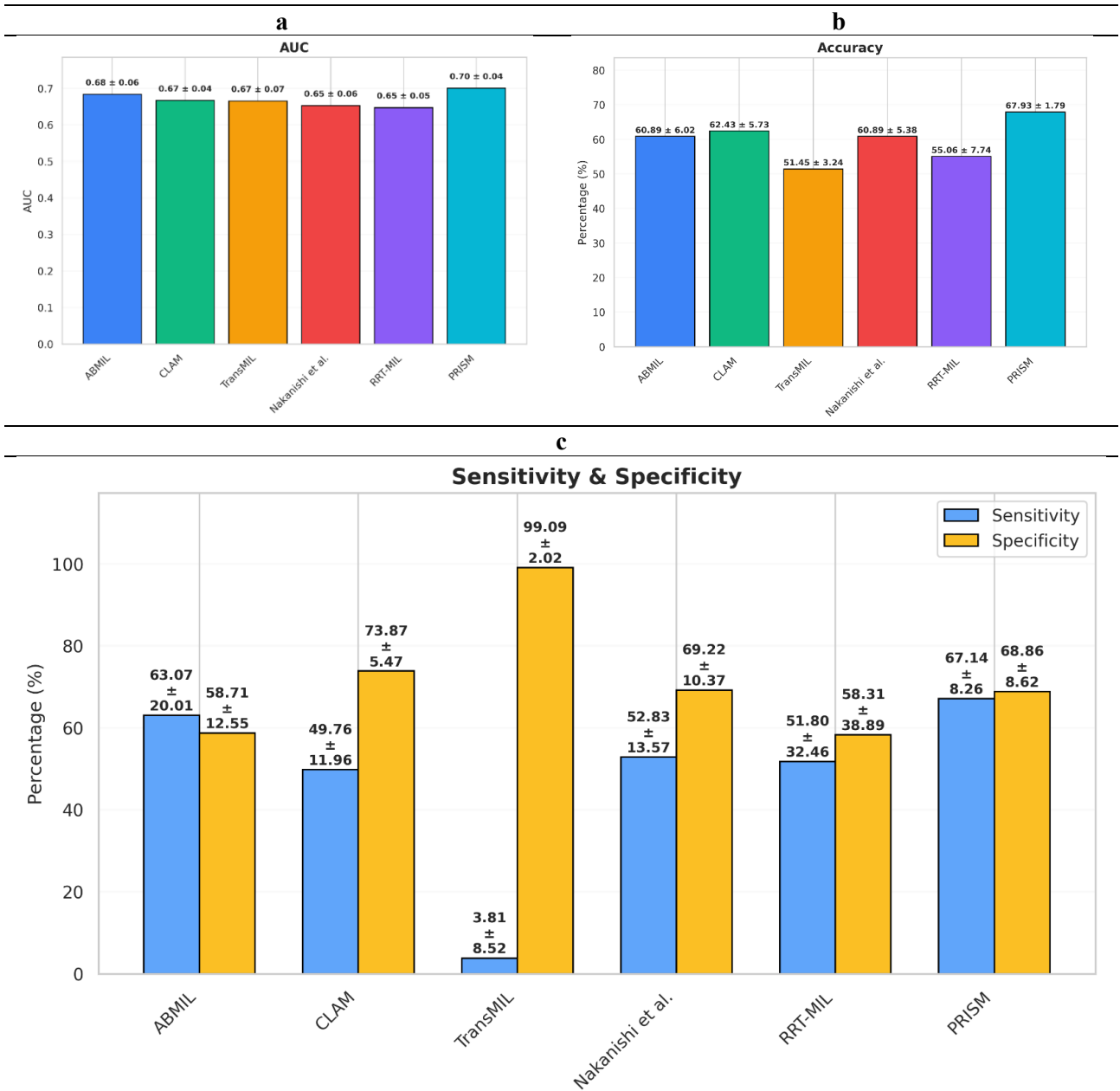

Figure S2: Five-year survival prediction results in the Alliance cohort using our proposed five-fold cross-validation method. (a) Area Under the Curve (AUC) values with standard deviations, (b) Accuracy percentages with standard deviations, and (c) Grouped comparison of sensitivity and specificity with other models. Our model achieves the highest accuracy and balanced sensitivity and specificity, demonstrating superior performance compared to existing state-of-the-art methods.

Table S4: Five-year survival prediction results of PRISM stratified by FL/IFL treatments in the Alliance cohort using our proposed validation strategy. Each row reports the average performance with standard deviation. For each metric within a treatment, the best result is shown in bold, and the second best is underlined.

| Model | Treatment | AUC | Accuracy (%) | Sensitivity (%) | Specificity (%) |
|---|---|---|---|---|---|
| ABMIL [30] | FL | $0.7118 \pm 0.058$ | $63.86 \pm 6.30$ | $65.26 \pm 17.11$ | $62.46 \pm 12.28$ |
|  | IFL | $0.6576 \pm 0.141$ | $58.00 \pm 10.85$ | $61.76 \pm 27.66$ | $54.24 \pm 15.40$ |
| CLAM [26] | FL | $0.6820 \pm 0.044$ | $63.29 \pm 4.45$ | $48.63 \pm 9.40$ | $77.96 \pm 7.96$ |

| | IFL | 0.6444 ± 0.112 | 59.85 ± 11.17 | 49.62 ± 22.04 | 70.09 ± 3.92 |
| --- | --- | --- | --- | --- | --- |
| Transmil [31] | FL | 0.7132 ± 0.057 | 51.89 ± 3.70 | 4.40 ± 8.80 | 99.35 ± 1.29 |
| | IFL | 0.6091 ± 0.098 | 51.09 ± 2.10 | 3.30 ± 6.60 | 98.80 ± 2.20 |
| Nakanishi et. Al [33] | FL | 0.6914 ± 0.080 | 62.21 ± 4.70 | 56.24 ± 17.12 | 68.18 ± 11.43 |
| | IFL | 0.6843 ± 0.128 | 58.83 ± 8.60 | 48.16 ± 15.27 | 69.51 ± 15.43 |
| RRT-MIL [32] | FL | 0.6961 ± 0.036 | 59.51 ± 11.11 | 59.82 ± 32.59 | 59.19 ± 36.46 |
| | IFL | 0.5962 ± 0.159 | 51.18 ± 9.42 | 46.30 ± 33.37 | 56.07 ± 42.83 |
| PRISM | FL | 0.7160 ± 0.061 | 68.21 ± 3.40 | 64.82 ± 3.40 | 71.60 ± 8.50 |
| | IFL | 0.6846 ± 0.108 | 66.77 ± 5.10 | 68.76 ± 13.90 | 64.77 ± 10.90 |

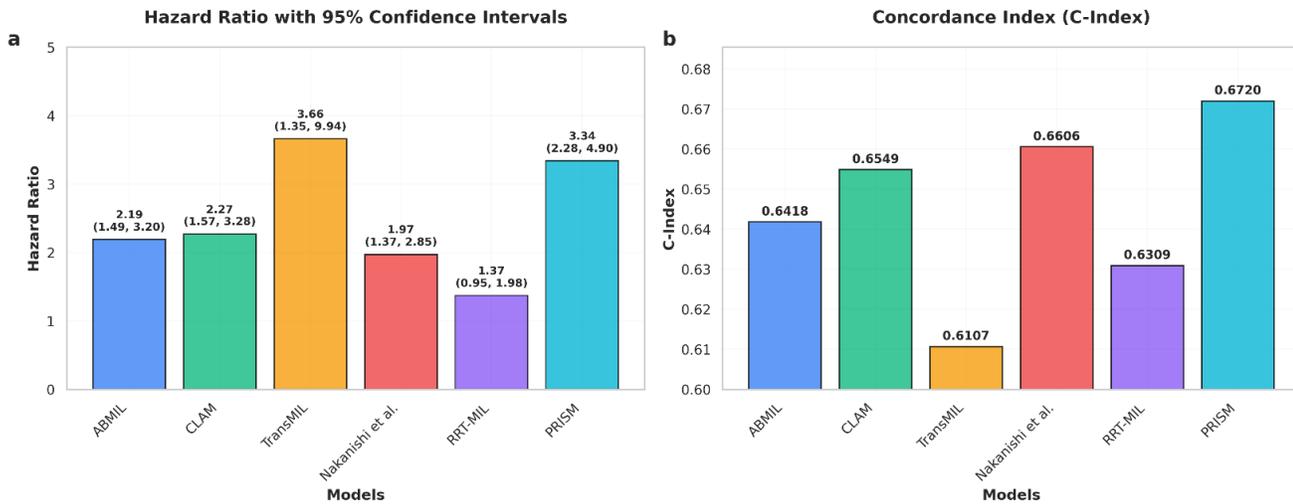

Figure S3: Hazard ratios and concordance index (c-index) values in the Alliance cohort using our proposed validation strategy. (a) Hazard ratios with 95% confidence intervals for each model, and (b) Concordance Index (C-Index) values demonstrating model ability to sort the patient based on risk for survival prediction. Our model achieves the highest C-Index (0.6720) and comparable hazard ratio in comparison with TransMIL with smaller confidence interval, indicating superior prognostic performance compared to other state-of-the-art methods.

Table S5: Five-year survival prediction results of PRISM stratified by sex in the Alliance cohort using our proposed validation strategy. Each row reports the average performance with standard deviation. For each metric, the best result is shown in bold, and the second-best is underlined.

| Model | Sex | AUC | Accuracy (%) | Sensitivity (%) | Specificity (%) |
| --- | --- | --- | --- | --- | --- |
| ABMIL [30] | Female | 0.6816 ± 0.079 | 61.71 ± 8.20 | 60.87 ± 19.10 | 62.54 ± 13.14 |
| | Male | 0.6906 ± 0.064 | 59.26 ± 5.10 | 64.51 ± 22.47 | 54.73 ± 15.35 |
| CLAM [26] | Female | 0.7046 ± 0.070 | 65.10 ± 8.70 | 56.34 ± 16.78 | 73.85 ± 8.72 |
| | Male | 0.6377 ± 0.053 | 59.90 ± 5.50 | 45.87 ± 13.08 | 73.93 ± 5.80 |
| Transmil [31] | Female | 0.6699 ± 0.110 | 52.85 ± 5.70 | 5.70 ± 11.42 | 1.0 ± 0.0 |
| | Male | 0.6541 ± 0.054 | 50.57 ± 1.10 | 2.80 ± 5.70 | 98.28 ± 3.40 |
| Nakanishi et. Al [33] | Female | 0.6796 ± 0.080 | 59.33 ± 6.20 | 46.70 ± 9.60 | 71.96 ± 14.18 |
| | Male | 0.6672 ± 0.043 | 59.96 ± 2.60 | 54.05 ± 14.40 | 65.86 ± 13.73 |
| RRT-MIL [32] | Female | 0.6299 ± 0.055 | 55.95 ± 7.08 | 52.46 ± 33.69 | 59.45 ± 37.99 |
| | Male | 0.6535 ± 0.056 | 54.37 ± 8.67 | 51.38 ± 31.69 | 57.36 ± 40.05 |
| PRISM | Female | 0.7120 ± 0.07 | 67.91 ± 3.20 | 64.00 ± 11.0 | 71.81 ± 10.00 |
| | Male | 0.6869 ± 0.060 | 68.06 ± 5.00 | 69.7 ± 9.00 | 66.38 ± 11.00 |

**Table S6:** c-index values got using TCGA-COADREAD dataset. All methods were trained using five-fold cross validation

| Model | C-index |
|---|---|
| ABMIL | 0.6034 ± 0.0500 |
| PANTHER | 0.5832 ± 0.0735 |
| DSMIL | 0.5 ± 0.0 |
| RRT-MIL | 0.5991 ± 0.08094 |
| **PRISM** | **0.6268 ± 0.0612** |